\documentclass{article}
\usepackage{spconf,amsmath,graphicx}
\usepackage{makecell}
\usepackage{multirow}
\usepackage{subcaption}
\usepackage{booktabs}
\usepackage{enumitem,xcolor}
\usepackage{tikz}
\usetikzlibrary{automata, arrows.meta, decorations.pathreplacing, positioning}
\usepackage{pgfplots}
\usepackage[hidelinks]{hyperref}

\pgfplotsset{compat=1.18}
\setlist{nosep, leftmargin=14pt}

\title{AI generalisation gap in comorbid sleep disorder staging}

\name{%
\begin{tabular}{@{}c@{}}
Saswata Bose$^{\diamond *}$ \hspace{0.4em}
Suvadeep Maiti$^{\clubsuit\diamond *}$ \hspace{0.4em}
Shivam Kumar Sharma$^{\diamond *}$\\
Mythirayee S$^{\ddagger}$ \hspace{0.4em}
Tapabrata Chakraborti$^{\dagger}$ \hspace{0.4em}
Srijitesh Rajendran$^{\ddagger}$ \hspace{0.4em}
Raju S. Bapi$^{\diamond}$
\end{tabular}
}

\address{%
$^{\diamond}$ International Institute of Information Technology (IIIT-H), Hyderabad, India\\
$^{\dagger}$ Alan Turing Institute, London and University College London, United Kingdom\\
$^{\clubsuit}$ Queen Square Institute of Neurology, University College London, United Kingdom\\
$^{\ddagger}$ National Institute of Mental Health and Neuroscience (NIMHANS), Bangalore, India
}

\begin{document}
\ninept
\maketitle
\begin{abstract}
Accurate sleep staging is essential for diagnosing OSA and hypopnea in stroke patients. Although PSG is reliable, it is costly, labor-intensive, and manually scored. While deep learning enables automated EEG-based sleep staging in healthy subjects, our analysis shows poor generalization to clinical populations with disrupted sleep. Using Grad-CAM interpretations, we systematically demonstrate this limitation. We introduce iSLEEPS, a newly clinically annotated ischemic stroke dataset (to be publicly released), and evaluate a SE-ResNet plus bidirectional LSTM model for single-channel EEG sleep staging. As expected, cross-domain performance between healthy and diseased subjects is poor. Attention visualizations, supported by clinical expert feedback, show the model focuses on physiologically uninformative EEG regions in patient data. Statistical and computational analyses further confirm significant sleep architecture differences between healthy and ischemic stroke cohorts, highlighting the need for subject-aware or disease-specific models with clinical validation before deployment. A summary of the paper and the code is available at \url{https://himalayansaswatabose.github.io/iSLEEPS_Explainability.github.io/}
\end{abstract}

\section{Introduction}
\label{sec:intro}
Sleep staging is critical for diagnosing and managing sleep disorders such as obstructive sleep apnea (OSA) and hypopnea, which disrupt sleep and degrade cognition and quality of life~\cite{redline2010obstructive}. Polysomnography (PSG), the clinical gold standard, uses EEG, EOG, and EMG to classify Wake, REM, and non-REM (N1–N3) stages, but it is costly, time-consuming, and prone to interscorer variability, especially for lighter stages like N1. Deep learning approaches provide automated and consistent alternatives with strong EEG-based performance~\cite{NAS, DeepSleepNet}, yet they are trained almost exclusively on healthy data, leaving their generalization to clinical populations with disrupted sleep architecture unclear. This issue is exacerbated by the scarcity of labeled pathological data~\cite{chen2019automatic}, highlighting the need for population-representative models and for understanding model behavior under pathology. Although sleep stage definitions should be health independent, degraded clinical performance suggests reliance on pathology-influenced or noisy features. Prior work has not modeled or interpreted sleep architecture in stroke patients with multiple comorbid sleep disorders, a group marked by severe neurological and electrophysiological abnormalities, including epilepsy, altered thalamocortical coupling, and asymmetric cortical activity, which make healthy-trained models unreliable and clinically misleading~\cite{ferre2013strokes}. The lack of public pathology-specific EEG datasets and tailored models creates a critical gap between computational advances and clinical practice. We bridge this gap by presenting the first stroke-specific sleep staging framework, combining deep learning with medically interpretable analysis and a newly curated PSG dataset of stroke patients with comorbid sleep disorders, enabling pathology-aware and clinically deployable sleep assessment for neurologically impaired populations.

\textbf{The contributions of this paper are:} 1. Introduced iSLEEPS (which is accessible at \url{https://tinyurl.com/iSLEEPSv1}), a new PSG dataset of 100 ischemic stroke patients with severe sleep disorders. 2. Developed a benchmarking deep learning model for iSLEEPS that achieves state-of-the-art performance and reveals the limited generalization of models trained on healthy cohorts. 3. Statistical and explainability analyses, validated by clinician feedback, showed deep models do not generalise between healthy and comorbid cohorts, underscoring the need for subject-aware or disease-specific sleep staging models.

\section{Methodology}
\label{sec:methods}
\subsection{Model Architecture}

\begin{figure}
	\centering
		\includegraphics[scale=0.25]{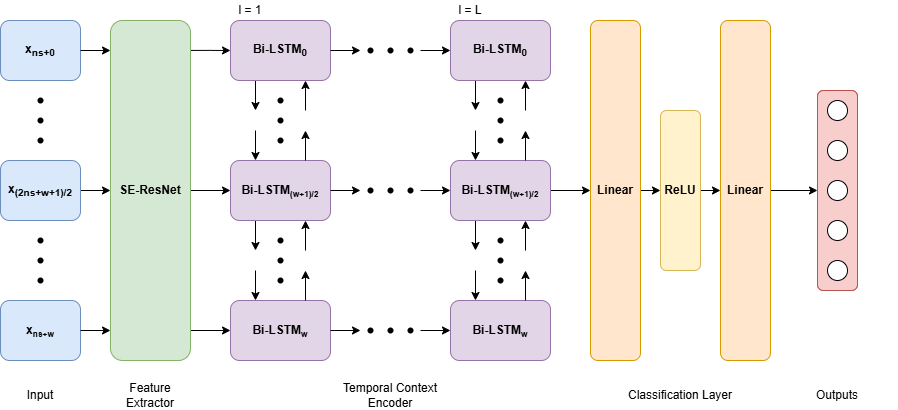}
	\caption{Deep Model Architecture, with the $n$\textsuperscript{th} epoch being input to the model, having a window size of $w$ and a stride length of $s$.}
	\label{FIG:1}
\end{figure}

The model (Fig. \ref{FIG:1}) uses a sliding window of consecutive 30 s EEG epochs to capture temporal context and predicts the sleep stage of the central epoch. A SE-ResNet block extracts discriminative spectral–temporal features by enhancing relevant frequency–amplitude patterns while suppressing noise, while stacked Bi-LSTM layers model bidirectional temporal dependencies and long-range stage transitions. The encoded features are then passed through fully connected layers, with the final layer producing sleep stage class probabilities.

\subsection{Explainability Methods}
We used GradCAM~\cite{gradcam} for explainability. Gradient-weighted Class Activation Mapping (Grad-CAM) generates class-specific heatmaps from the gradients of a target class with respect to the final convolutional layer. In signal classification, it highlights time or frequency regions most influential to the model’s decisions, enabling assessment of whether the model attends to meaningful physiological features (e.g., sleep spindles or alpha rhythms), thereby improving interpretability, supporting debugging, and ensuring alignment with domain knowledge.

\section{Experiments}
\label{sec:exp}

We evaluated performance using raw single-channel EEG (C4:M1) from SleepEDF-20~\cite{sleepedf}, SleepEDF-78, SHHS~\cite{shhs}, and our iSLEEPS dataset. SleepEDF-20 includes 39 PSG recordings from 20 healthy adults aged 25–34 with R\&K staging (W, N1, N2, N3 [N3+N4], REM), excluding Movement and UNKNOWN stages. SleepEDF-78 extends this to 78 participants. SHHS provides full-night PSG data from 6,441 subjects scored under AASM guidelines, with analysis restricted to 329 typical sleepers. iSLEEPS comprises 100 ischemic stroke patients (mean age 50.52; 23 females, 77 males) recruited at NIMHANS, Bangalore, India. It shows a high rate of sleep apnea (38\% severe, 23\% moderate) with detailed obstructive/central apnea annotations Deep learning models were trained separately per dataset using Adam optimizer(learning rate 0.001), negative log-likelihood loss, batch size 128, and evaluated via leave-one-out cross-validation (20, 10, 5, and 10 folds for SleepEDF-20, SleepEDF-78, SHHS, and iSLEEPS, respectively). The architecture processes raw EEG with a window size 9, stride 4, and three stacked LSTM layers, with the best healthy-subject model tested on iSLEEPS to simulate clinical deployment, followed by GradCAM analysis on five randomly selected patients to assess errors and alignment with clinically relevant EEG patterns.

\section{Results and Discussion}
\subsection{Model Performance}

\begin{table}
\resizebox{0.5\textwidth}{!}{
\begin{tabular}{llccc}
\hline
\multirow{2}{*}{\textbf{Datasets}} & \multirow{2}{*}{\textbf{Model}} & \multicolumn{3}{c}{\textbf{Overall Results}}  \\ \cline{3-5} 
                                   &                                 & \textbf{ACC}  & \textbf{MF1}  & \textbf{$\kappa$}    \\ \hline
\multirow{5}{*}{\makecell[l]{SleepEDF-20 \\ (Healthy Cohort)}} & NAS~\cite{NAS}                             & 82.7          & 75.9          & 0.76          \\
                                   & DeepSleepNet~\cite{DeepSleepNet}                    & 82.0          & 76.9          & 0.76          \\
                                   & XSleepNet2~\cite{XSleepNet}                      & 83.9          & 78.7          & 0.77          \\
                                   & SleepContextNet~\cite{SleepContextNet}                 & 84.8          & 79.8          & 0.79          \\
                                   & SEResnet-Transformer~\cite{10446703}                 & 79.3          & 74.7          & 0.72          \\
                                   & \textbf{Ours}                   & \textbf{87.5} & \textbf{82.5} & \textbf{0.82} \\ \hline
\multirow{4}{*}{\makecell[l]{SleepEDF-78 \\ (Healthy Cohort)}} & NAS~\cite{NAS}                             & 80.0          & 72.7          & 0.72          \\
                                   & XSleepNet2~\cite{XSleepNet}                      & 80.3          & 76.4          & 0.73          \\
                                   & SleepContextNet~\cite{SleepContextNet}                 & 82.7          & 77.2          & 0.76          \\
                                   & SEResnet-Transformer~\cite{10446703}                 & 73.6          & 70.6          & 0.68          \\
                                   & \textbf{Ours}                   & \textbf{83.8} & \textbf{78.9} & \textbf{0.77} \\ \hline
\multirow{3}{*}{\makecell[l]{SHHS \\ (Healthy Cohort)}} & NAS~\cite{NAS}                             & 81.9          & 75.3          & 0.74          \\
                                   & SleepContextNet~\cite{SleepContextNet}                 & 86.4          & 80.5          & 0.81          \\
                                   & SEResnet-Transformer~\cite{10446703}                 & 79.0          & 69.3          & 0.71          \\
                                   & \textbf{Ours}                   & \textbf{87.8} & \textbf{81.9} & \textbf{0.83} \\ \hline
\multirow{3}{*}{\makecell[l]{iSLEEPS \\ (Patient Cohort)}} & ResNet~\cite{ResNet}                       & 61.6          & 54.4          & 0.48          \\
                                   & SEResnet-Transformer~\cite{10446703}                 & 67.4          & 59.3          & 0.54          \\
                                   & \textbf{Ours}                   & \textbf{74.7} & \textbf{67.7} & \textbf{0.64} \\ \hline
\end{tabular}
}
\caption{Results comparison with different models, trained and tested on SleepEDF-20, SleepEDF-78, SHHS and iSLEEPS dataset, respectively. Performance of our model is shown in bold. The performance has been measured using Accuracy (ACC), Macro F1 Score (MF1) and Kappa Score ($\kappa$), which shows that the model predictions align substantially with respect to random chance.}
\label{tab:TAB1}
\end{table}
We provide a comprehensive comparison with the current literature in Table \ref{tab:TAB1} and an overview of the model's sleep stage-wise performance across four different datasets in Table \ref{tab:TAB2}. This validates the fact the proposed benchmarking model on iSLEEPS is at par and, in fact, better than the existing models in the current literature for sleep staging in the available healthy subject datasets, as well as in patient-based cohorts.

\begin{table}
\centering
\renewcommand{\arraystretch}{1.1}
\setlength{\tabcolsep}{4pt}
\begin{tabular}{lccccc}
\toprule
\textbf{Dataset} & \textbf{W} & \textbf{N1} & \textbf{N2} & \textbf{N3} & \textbf{REM} \\ 
\midrule
SleepEDF-20 & 92.0 & 56.9 & 89.9 & 87.9 & 85.9 \\
SleepEDF-78 & 92.1 & 50.1 & 85.2 & 82.1 & 82.1 \\
SHHS        & 89.1 & 54.4 & 89.2 & 87.9 & 88.9 \\
iSLEEPS     & 79.9 & 32.9 & 80.9 & 74.2 & 70.6 \\
\bottomrule
\end{tabular}
\caption{Sleep stage classification (using single-channel raw EEG input) class-wise performance (F1 Score) of our model on the healthy subject datasets: SleepEDF-20, SleepEDF-78, and SHHS; and the patient dataset, iSLEEPS.}
\label{tab:TAB2}
\end{table}

\begin{figure}
    \centering
    \begin{subfigure}{1\linewidth}
        \centering
        \includegraphics[width=1\linewidth]{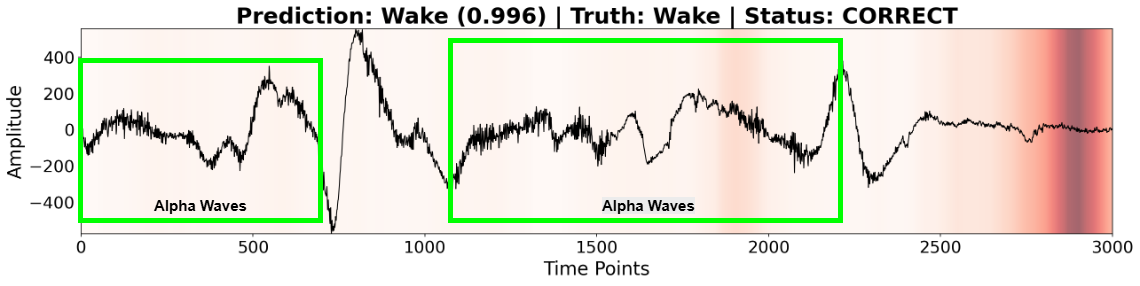}
        \caption{A Correctly Tagged Epoch. The green box highlights the Alpha Waves present in the epoch.}
        \label{fig:sub_1}
    \end{subfigure}
    
    \begin{subfigure}{1\linewidth}
        \centering
        \includegraphics[width=1\linewidth]{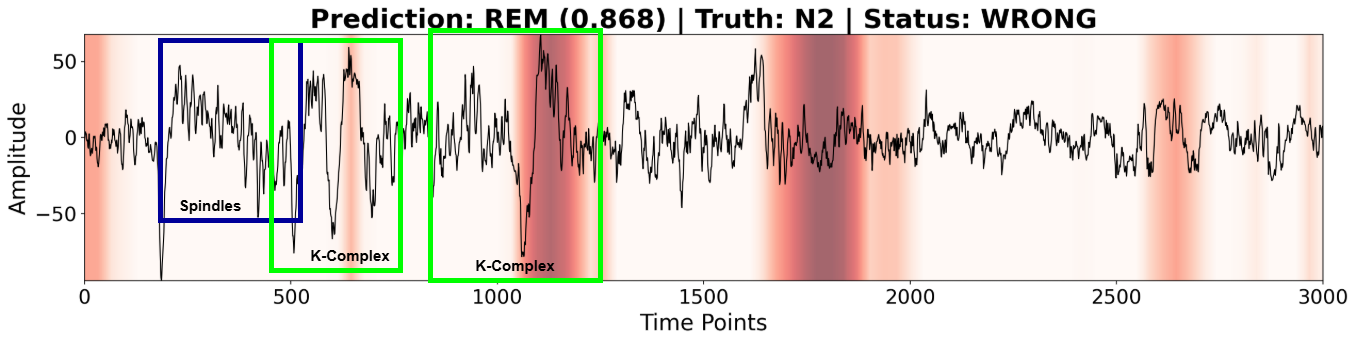}
        \caption{An Incorrectly Tagged Epoch. The green box highlights the K-Complexes and the blue box highlights the Sleep Spindles present in the epoch.}
        \label{fig:sub_2}
    \end{subfigure}
    \caption{Medically Annotated Epochs from iSLEEPS showing well-focussed and ill-focussed sleep-relevant artefacts based on the model trained on SHHS and tested on iSLEEPS. The title also includes the probability of the predicted class provided by the model in brackets.}
    \label{fig:combined}
\end{figure}

\begin{figure}
    \centering
    \includegraphics[width=1\linewidth]{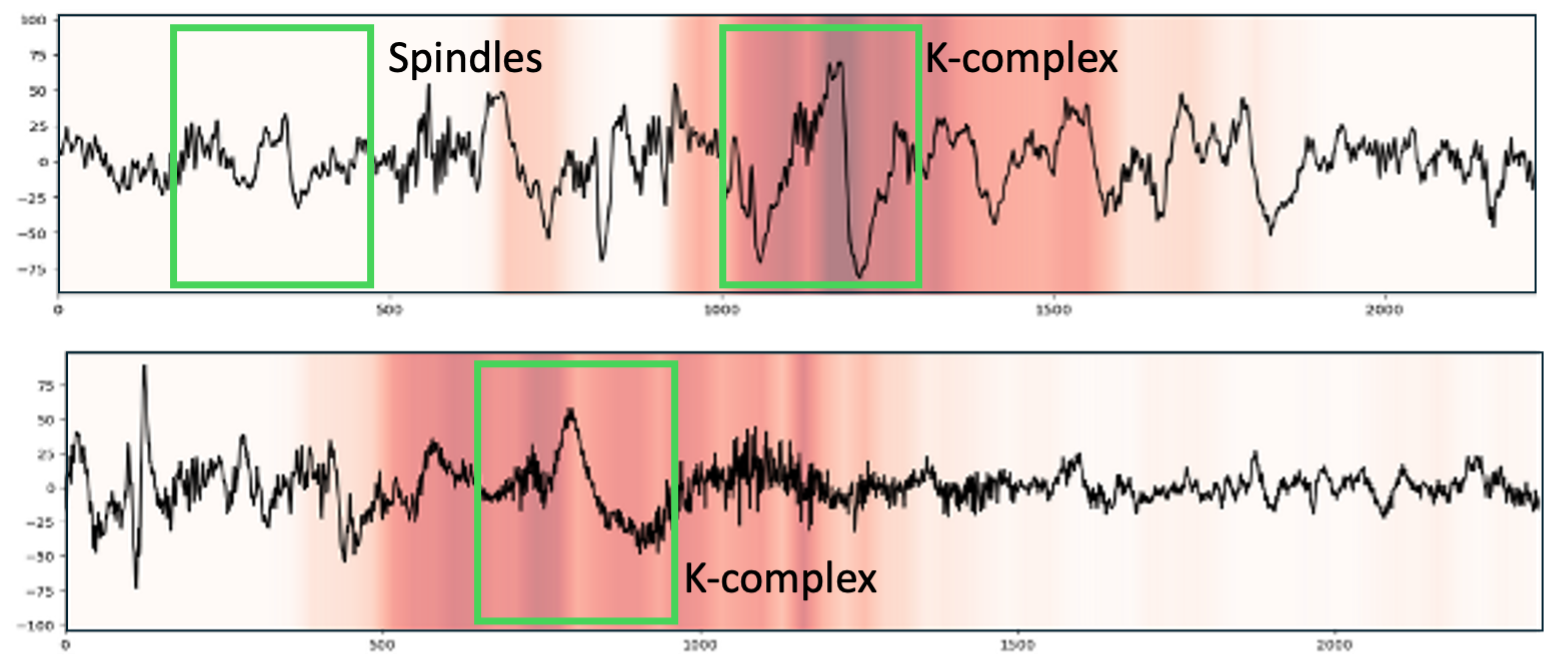}
    \caption{GradCAM visualization of raw EEG epoch of sleep stage N2 (correctly predicted by the model) and boxes in green indicating K-complex and spindles, extracted from the model trained and tested on iSLEEPS.}
    \label{fig:patient_n2}
\end{figure}

\subsection{Ablation studies}
We performed ablation studies on the SleepEDF-20 and iSLEEPS datasets using 20-fold and 10-fold cross-validation, respectively, to assess the impact of key model components. A window size of 9 with a stride of 4 yielded the best balance between temporal coverage and efficiency. SE-ResNet-18 outperformed SE-ResNet-34 as the feature extractor, with the latter showing overfitting, and the SE module significantly improved feature learning. Optimal performance was achieved with SE-ResNet-18 and three Bi-LSTM stacks, as additional stacks increased complexity without improving results.

\subsection{Model Explainability and Clinician Feedback}
For the 100 iSLEEPS patients, sleep stages were evaluated using the SHHS-trained model, yielding 55.1\% accuracy and 51.9\% MF1, a sharp drop from Table \ref{tab:TAB1}. To interpret its behavior, GradCAM heatmaps were generated for all epochs of four randomly selected patients. Medical experts found that, despite training on healthy data, the model often attended to physiologically irrelevant EEG regions in stroke patients. Although some epochs showed clinically meaningful patterns, many activations reflected diffuse slowing, intermittent delta bursts, or movement and electrode artifacts. They did not match typical sleep markers such as spindles, K-complexes, or alpha attenuation. This mismatch was most pronounced during N1–N2 transitions and REM detection, where attention favored ischemia-related slowing or hemispheric asymmetries over lesioned areas.

Overall, the model generalizes poorly from healthy subjects to patients, frequently misdirecting attention, missing clinically relevant cues, and producing systematic staging errors. Figure \ref{fig:combined} examines two representative epochs (trained on healthy data, tested on patients) using medical annotations and GradCAM. In Figure \ref{fig:sub_1}, the prediction is correct but attention is concentrated on physiologically irrelevant regions, indicating poor medical alignment. In Figure \ref{fig:sub_2}, attention is closer to relevant regions, yet the prediction is incorrect, again highlighting a disconnect between clinical reasoning and model behavior. By contrast, a model trained and evaluated on iSLEEPS (Table \ref{tab:TAB1}) shows improved focus on clinically meaningful features (Figure \ref{fig:patient_n2}), suggesting superior capture of patient-specific sleep architecture.

These results indicate that models trained on healthy-subject datasets (SleepEDF-20, SleepEDF-78, SHHS) should be applied in clinical or diagnostic settings only under medical supervision. This limitation is expected, as sleep architecture differs substantially between ischemic stroke patients and healthy individuals \cite{ferre2013strokes}, a disparity our experiments explicitly expose through the observed performance degradation.

\subsection{Clincian-in-the-loop Analysis of Sleep Transition Graphs}
We isolated epochs from SleepEDF-78 and iSLEEPS to form a balanced dataset containing 22,430 (49\%) epochs from SleepEDF-78 and 23,192 (51\%) from iSLEEPS, ensuring substantial representation of both healthy and patient samples. Significance testing confirmed that the sleep architectures of the two groups differ notably in stage frequency and order. The state-transition diagrams (Fig \ref{fig:sleep_side_by_side}) also show significant differences (diagram excludes insignificant transitions and self-transitions), with an average KL Divergence of $\approx0.17$, a Chi-Square Statistic of $\approx1831.4552$ with 19 degrees of freedom, and p-values ranging from $\approx1.33\times10^{(-37)}$ to $\approx1.87\times10^{(-2)}$ across 13 significant transitions out of 20 (Mean p-value $\approx 4.83\times10^{(-3)}$, Median p-value $\approx 2.24\times10^{(-6)}$). The analysis in this section was performed in direct collaboration with our clinician co-author.




\tikzset{
  >=Stealth,
  state/.style={
    circle, draw, thick, minimum size=8mm, inner sep=0pt, font=\scriptsize
  },
  wstate/.style={state, fill=yellow!20},
  nstate/.style={state, fill=blue!10},
  rstate/.style={state, fill=red!10},
  edgep/.style={->, line cap=round},
  elabel/.style={fill=white, inner sep=1pt, font=\scriptsize}
}
\newcommand{\probcolor}[1]{%
  blue!#1!red!100%
}

\newcommand{\valtocolor}[1]{%
  \pgfmathsetmacro{\t}{(#1 + 1)/2}%
  \pgfmathsetmacro{\t}{max(0, min(1, \t))}%
  \pgfmathsetmacro{\R}{(\t <= 0.5) ? (2*\t) : (1)}%
  \pgfmathsetmacro{\G}{(\t <= 0.5) ? (2*\t) : (2 - 2*\t)}%
  \pgfmathsetmacro{\B}{(\t <= 0.5) ? (1 - 2*\t) : (0)}%
  \definecolor{edgevalcolor}{rgb}{\R,\G,\B}%
}

\newcommand{\drawvaledge}[5]{%
  \pgfmathsetmacro{\p}{#3}%
  \pgfmathsetmacro{\v}{2*\p - 1}%
  \valtocolor{\v}%
  \draw[edgep, bend left=#4, line width = 0.8 pt, draw=edgevalcolor] 
    (#1) to node[elabel, pos=#5]{\scriptsize #3} (#2);%
}

\begin{figure}[htb]
  \centering

  \begin{subfigure}[t]{\columnwidth}
    \centering
    \begin{tikzpicture}[scale=0.90, every node/.style={scale=0.90}]
      \def\W{80} 
      \def\H{4}  
      \foreach \i in {0,...,160} {%
        \pgfmathsetmacro{\xmin}{-0.5*\W + \W*\i/161}%
        \pgfmathsetmacro{\xmax}{-0.5*\W + \W*(\i+1)/161}%
        \pgfmathsetmacro{\v}{-1 + 2*(\i/161)}%
        \valtocolor{\v}%
        \fill[edgevalcolor] (\xmin mm, -0.5*\H mm) rectangle (\xmax mm, 0.5*\H mm);%
      }%
      \draw[thick] (-0.5*\W mm, -0.5*\H mm) rectangle (0.5*\W mm, 0.5*\H mm);

      \foreach \v/\x in {0.0/-0.5*\W, 0.25/-0.25*\W, 0.5/0, 0.75/0.25*\W, 1.0/0.5*\W} {%
        \draw[thick] (\x mm, -0.5*\H mm) -- (\x mm, -0.5*\H mm - 1.5mm);
        \node[below, font=\scriptsize] at (\x mm, -0.5*\H mm - 1.5mm) {\v};
      }
      \node[above, font=\small] at (0, 0.5*\H mm) {Transition Probability};
    \end{tikzpicture}
  \end{subfigure}

  \vspace{0.7ex} 
  \begin{subfigure}[t]{0.45\columnwidth} 
    \centering
    \begin{tikzpicture}[scale=0.70, every node/.style={scale=0.80}]
      \node[wstate] (Wi)   at (90:25mm)  {W};
      \node[rstate] (REMi) at (162:25mm) {REM};
      \node[nstate] (N3i)  at (234:25mm) {N3};
      \node[nstate] (N2i)  at (306:25mm) {N2};
      \node[nstate] (N1i)  at (18:25mm)  {N1};

      \drawvaledge{N1i}{REMi}{0.068}{12}{0.6}
      \drawvaledge{N2i}{N1i}{0.223}{-18}{0.5}
      \drawvaledge{N2i}{N3i}{0.242}{18}{0.5}
      \drawvaledge{N2i}{REMi}{0.181}{12}{0.8}
      \drawvaledge{N2i}{Wi}{0.354}{10}{0.5}
      \drawvaledge{N3i}{N1i}{0.080}{12}{0.5}
      \drawvaledge{N3i}{N2i}{0.699}{12}{0.5}
      \drawvaledge{N3i}{REMi}{0.058}{18}{0.5}
      \drawvaledge{N3i}{Wi}{0.163}{8}{0.15}
      \drawvaledge{REMi}{N1i}{0.263}{12}{0.45}
      \drawvaledge{REMi}{N2i}{0.415}{12}{0.4} 
      \drawvaledge{Wi}{N1i}{0.629}{12}{0.5}
      \drawvaledge{Wi}{N2i}{0.344}{12}{0.7}
    \end{tikzpicture}
    \caption{iSLEEPS Transition Graph}
    \label{fig:isleeps}
  \end{subfigure}
  \hfill
  \begin{subfigure}[t]{0.45\columnwidth} 

    \begin{tikzpicture}[scale=0.70, every node/.style={scale=0.80}]
      \node[wstate] (Wc)    at (90:25mm)  {W};
      \node[rstate] (REMc)  at (162:25mm) {REM};
      \node[nstate] (N3c)   at (234:25mm) {N3};
      \node[nstate] (N2c)   at (306:25mm) {N2};
      \node[nstate] (N1c)   at (18:25mm)  {N1};

      \drawvaledge{N1c}{REMc}{0.104}{12}{0.6}
      \drawvaledge{N2c}{N1c}{0.289}{-18}{0.5}
      \drawvaledge{N2c}{N3c}{0.418}{18}{0.5}
      \drawvaledge{N2c}{REMc}{0.130}{12}{0.8}
      \drawvaledge{N2c}{Wc}{0.164}{10}{0.5}
      \drawvaledge{N3c}{N1c}{0.028}{12}{0.5}
      \drawvaledge{N3c}{N2c}{0.918}{12}{0.5}
      \drawvaledge{N3c}{REMc}{0.005}{18}{0.5}
      \drawvaledge{N3c}{Wc}{0.049}{8}{0.15}
      \drawvaledge{REMc}{N1c}{0.469}{12}{0.45}
      \drawvaledge{REMc}{N2c}{0.222}{12}{0.4}
      \drawvaledge{Wc}{N1c}{0.933}{12}{0.5}
      \drawvaledge{Wc}{N2c}{0.044}{12}{0.7}
    \end{tikzpicture}
    \caption{SleepEDF-78 Transition Graph}
    \label{fig:sleepcassette}
  \end{subfigure}

  \caption{Transition graph of the dataset (excluding insignificant transitions and self transitions) elucidating the probabilities of transition from one state to the other for the corresponding dataset. The nodes represent corresponding sleep stages}
  \label{fig:sleep_side_by_side}
\end{figure}
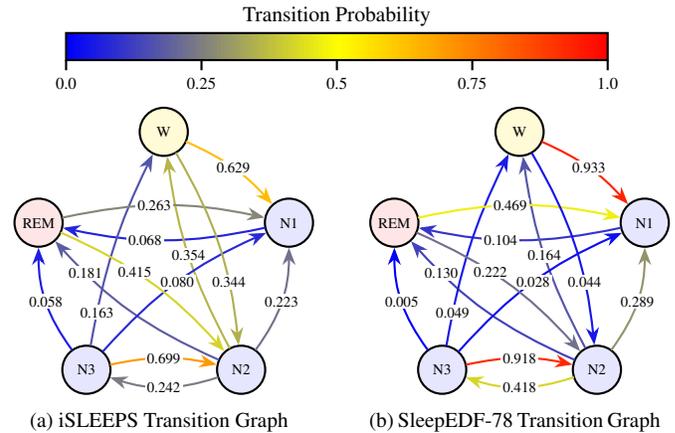

\textbf{The Statistical Lens:} Transition probability analysis revealed clear differences between stroke patients (iSLEEPS) and healthy individuals (SleepEDF-78). Stroke patients showed a higher likelihood of waking from light non-REM sleep (N2$\rightarrow$Wake: 0.354 vs. 0.164), indicating unstable sleep marked by frequent arousal rather than progression into deep restorative sleep. Clinically, this instability reduces sleep efficiency and restorative benefits, consistent with post-stroke impairment of thalamocortical and brainstem systems regulating sleep and arousal. Stroke patients showed slower N2$\rightarrow$N3 transitions, suggesting disrupted sleep depth. Healthy individuals showed stronger transitions, consistent with stable and consolidated sleep. These findings align with prior studies showing that cortical and subcortical brain lesions disrupt neural networks and lead to sleep disturbances after stroke~\cite{ferre2013strokes}.

\begin{figure}[]
    \centering
    \includegraphics[width=\linewidth]{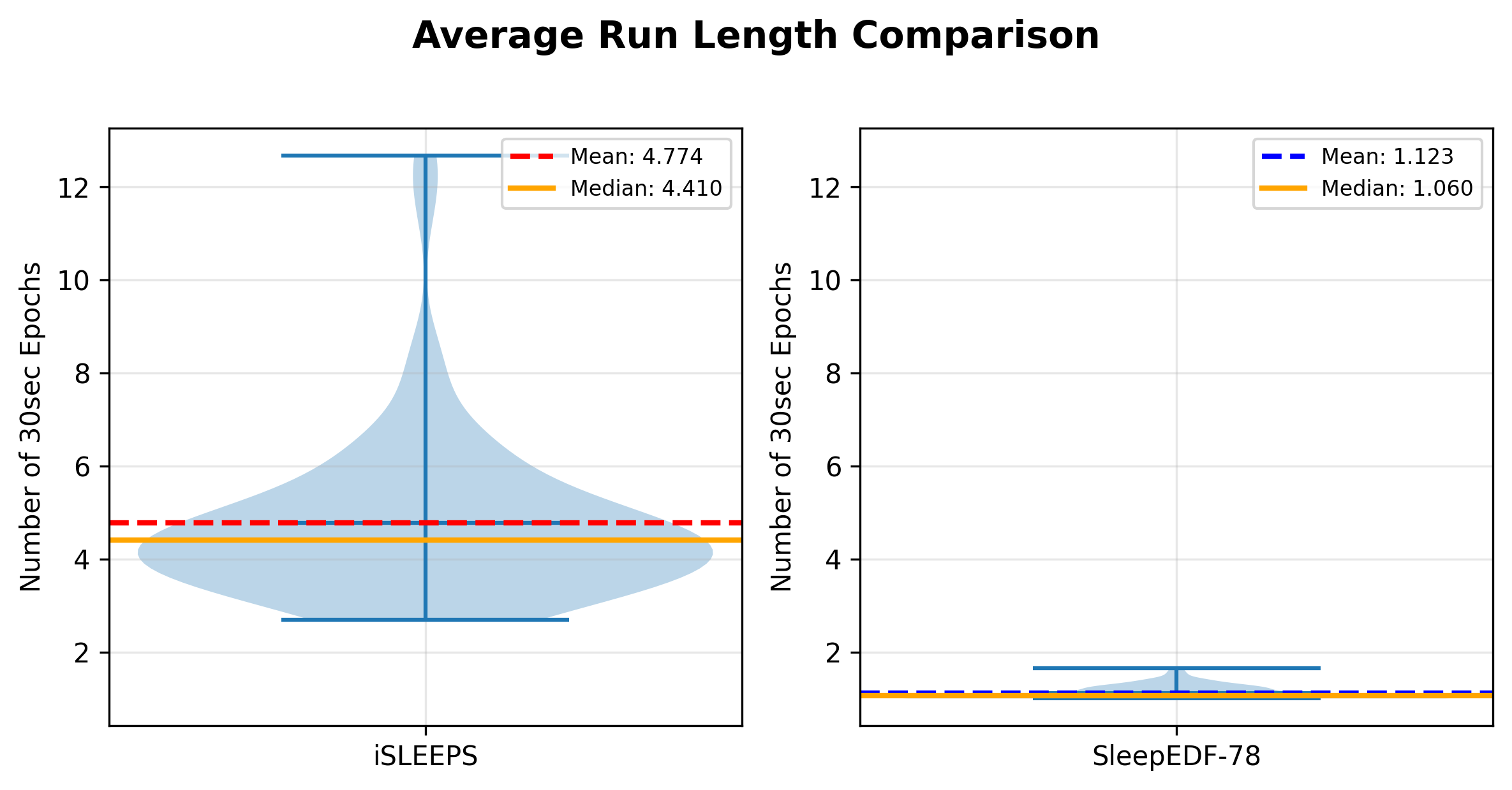}
    \caption{Distribution of the Average Run Length in the Datasets. The Violin Chart clearly highlights that the iSLEEPS dataset has a much wider range for the feature while SleepEDF-78 is more constrained within a smaller range of values.}
    \label{fig:avg_run_length_violin}
\end{figure}

\textbf{The Computational Lens:} Given clear differences between patient and healthy samples, we trained binary classifiers (Random Forest, Logistic Regression, Decision Tree, SVM) to predict whether an unepoched EEG recording came from a patient or a healthy subject. Using targeted feature engineering, we extracted 27 structural and sequential features; One of such features which perfectly separated the cohorts was Average Run Length (a \textbf{Run} is defined as a continuous segment without stage change) as seen in Figure \ref{fig:avg_run_length_violin}. Combinations of bigrams (e.g., Wake$\rightarrow$N1 with N3$\rightarrow$N2), Shannon entropy of the arrangement of the sleep stages, and maximum Run length also achieved 100\% accuracy with zero AUC-ROC variance across five folds. These findings show that ischemic stroke patients have markedly distinct sleep architecture, explaining poor generalization from healthy-only models and highlighting the need for domain-aware clinical sleep staging.

\section{Conclusion}
Models trained only on healthy sleep data perform poorly on patient sleep staging, showing an $\approx$30\% accuracy drop and strong dataset bias, and no public benchmarks exist for multimorbid sleep disorders. We highlight two urgent needs: models that handle inter-patient and structural variability, such as hierarchical approaches separating healthy and patient data, and mandatory expert review before clinical use due to reliance on non-pathological features. To address this gap, we introduce iSLEEPS, a forthcoming public benchmark, noting that medical-grade, explainable, and unbiased sleep staging remains an open challenge under high-risk Health AI and the EU AI Act~\cite{van2024eu}.

\bibliographystyle{IEEEbib}
\bibliography{strings,refs}

@article{sleepedf,
  title={Analysis of a sleep-dependent neuronal feedback loop: the slow-wave microcontinuity of the EEG},
  author={Kemp, Bob and Zwinderman, Aeilko H and Tuk, Bert and Kamphuisen, Hilbert AC and Oberye, Josefien JL},
  journal={IEEE Transactions on Biomedical Engineering},
  volume={47},
  number={9},
  pages={1185--1194},
  year={2000},
  publisher={IEEE}
}

@article{shhs,
  title={The National Sleep Research Resource: towards a sleep data commons},
  author={Zhang, Guo-Qiang and Cui, Licong and Mueller, Remo and Tao, Shiqiang and Kim, Matthew and Rueschman, Michael and Mariani, Sara and Mobley, Daniel and Redline, Susan},
  journal={Journal of the American Medical Informatics Association},
  volume={25},
  number={10},
  pages={1351--1358},
  year={2018},
  publisher={Oxford University Press}
}

@inproceedings{gradcam,
  title={Grad-cam: Visual explanations from deep networks via gradient-based localization},
  author={Selvaraju, Ramprasaath R and Cogswell, Michael and Das, Abhishek and Vedantam, Ramakrishna and Parikh, Devi and Batra, Dhruv},
  booktitle={Proceedings of the IEEE international conference on computer vision},
  pages={618--626},
  year={2017}
}

@article{chen2019automatic,
  title={Automatic sleep stage classification based on subthalamic local field potentials},
  author={Chen, Yuxuan and Gong, Chao and Hao, Haiming and Guo, Yan and Xu, Shuang and Zhang, Yan and Yin, Guangyu and Cao, Xingshi and Yang, Aiping and Meng, Fan and Ye, Jing and Liu, Haifeng and Zhang, Jian and Sui, Yuelin and Li, Luming},
  journal={IEEE Transactions on Neural Systems and Rehabilitation Engineering},
  volume={27},
  number={2},
  pages={118--128},
  year={2019},
  publisher={IEEE},
  doi={10.1109/TNSRE.2018.2890272},
  pmid={30605104},
  pmcid={PMC6544463}
}

@article{redline2010obstructive,
  title={Obstructive sleep apnea--hypopnea and incident stroke: the sleep heart health study},
  author={Redline, Susan and Yenokyan, Gayane and Gottlieb, Daniel J and Shahar, Eyal and O'Connor, George T and Resnick, Helaine E and Diener-West, Marie and Sanders, Mark H and Wolf, Philip A and Geraghty, Estella M and others},
  journal={American journal of respiratory and critical care medicine},
  volume={182},
  number={2},
  pages={269--277},
  year={2010},
  publisher={American Thoracic Society}
}

@ARTICLE{NAS,
  author={Kong, Gangwei and Li, Chang and Peng, Hu and Han, Zhihui and Qiao, Heyuan},
  journal={IEEE Transactions on Neural Systems and Rehabilitation Engineering}, 
  title={EEG-Based Sleep Stage Classification via Neural Architecture Search}, 
  year={2023},
  volume={31},
  number={},
  pages={1075-1085},
  doi={10.1109/TNSRE.2023.3238764}}

@ARTICLE{DeepSleepNet,
  author={Supratak, Akara and Dong, Hao and Wu, Chao and Guo, Yike},
  journal={IEEE Transactions on Neural Systems and Rehabilitation Engineering}, 
  title={DeepSleepNet: A Model for Automatic Sleep Stage Scoring Based on Raw Single-Channel EEG}, 
  year={2017},
  volume={25},
  number={11},
  pages={1998-2008}
}

@article{XSleepNet,
  title={XSleepNet: Multi-view sequential model for automatic sleep staging},
  author={Phan, Huy and Ch{\'e}n, Oliver Y and Tran, Minh C and Koch, Philipp and Mertins, Alfred and De Vos, Maarten},
  journal={IEEE Transactions on Pattern Analysis and Machine Intelligence},
  volume={44},
  number={9},
  pages={5903--5915},
  year={2021}
}

@article{SleepContextNet,
  title={SleepContextNet: A temporal context network for automatic sleep staging based single-channel EEG},
  author={Zhao, Caihong and Li, Jinbao and Guo, Yahong},
  journal={Computer Methods and Programs in Biomedicine},
  volume={220},
  pages={106806},
  year={2022},
  publisher={Elsevier}
}

@inproceedings{ResNet,
  title={Deep residual learning for image recognition},
  author={He, Kaiming and Zhang, Xiangyu and Ren, Shaoqing and Sun, Jian},
  booktitle={Proceedings of the IEEE conference on computer vision and pattern recognition},
  pages={770--778},
  year={2016}
}

@INPROCEEDINGS{10446703,
  author={Maiti, Suvadeep and Sharma, Shivam Kumar and Bapi, Raju S.},
  booktitle={ICASSP 2024 - 2024 IEEE International Conference on Acoustics, Speech and Signal Processing (ICASSP)}, 
  title={Enhancing Healthcare with EOG: A Novel Approach to Sleep Stage Classification}, 
  year={2024},
  volume={},
  number={},
  pages={2305-2309},
  doi={10.1109/ICASSP48485.2024.10446703}}

@article{ferre2013strokes,
  title={Strokes and their relationship with sleep and sleep disorders},
  author={Masó, Alex Ferré and others},
  journal={Neurologia (English Edition)},
  volume={28},
  number={2},
  pages={103--118},
  year={2013},
  publisher={Elsevier}
}

@article{van2024eu,
  title={The EU artificial intelligence act (2024): implications for healthcare},
  author={Van Kolfschooten, Hannah and Van Oirschot, Janneke},
  journal={Health Policy},
  volume={149},
  pages={105152},
  year={2024},
  publisher={Elsevier}
}

\section {Ethical Compliance}
The study was carried out at the NIMHANS, Bengaluru, India. Approval was obtained from the NIMHANS Institutional Ethics Committee [No. NIMHANS/34th IEC (BS\&NS DIV.)/2022 dated 05.02.2022]. Ethical considerations were strictly followed throughout the study.

\section{Acknowledgements}
We acknowledge IHub-Data, IIIT Hyderabad (H1-002), for financial assistance. Tapabrata Chakraborti is supported by the Turing-Roche Strategic Partnership and UCL NIHR Biomedical Research Centre.

\section{Conflict of Interest}
The authors declare no competing interests.

\end{document}